\documentclass[10pt,twocolumn,letterpaper]{article}

\usepackage{iccv}
\usepackage{times}
\usepackage{epsfig}
\usepackage{graphicx}
\usepackage{amsmath}
\usepackage{amssymb}
\usepackage{color}
\usepackage{subcaption}
\usepackage{enumitem}

\newcommand{\argmin}[1]{\underset{#1}{\operatorname{arg}\,\operatorname{min}}\;}

\usepackage[breaklinks=true,bookmarks=false]{hyperref}

\iccvfinalcopy 

\def\iccvPaperID{1366} 

\ificcvfinal\fi
\begin{document}
\raggedbottom 

\title{Fast Face-swap Using Convolutional Neural Networks}

\author{Iryna Korshunova$^{1,2}$
\and
Wenzhe Shi$^1$
\and
Joni Dambre$^2$
\and
Lucas Theis$^1$
\and
$^1$Twitter\\
{\tt\small \{iryna.korshunova, joni.dambre\}@ugent.be}
\and
$^2$IDLab, Ghent University\\
{\tt\small \{wshi, ltheis\}@twitter.com}
}

\maketitle
\thispagestyle{empty}

\begin{abstract}
	We consider the problem of face swapping in images, where an input identity is
	transformed into a target identity while preserving pose, facial expression and lighting. To
	perform this mapping, we use convolutional neural networks trained to capture the appearance of
	the target identity from an unstructured collection of his/her photographs. This approach is
	enabled by framing the face swapping problem in terms of style transfer, where the goal is to
	render an image in the style of another one. Building on recent advances in this area, we devise
	a new loss function that enables the network to produce highly photorealistic results. By
	combining neural networks with simple pre- and post-processing steps, we aim at making face swap
	work in real-time with no input from the user.
\end{abstract}

\section{Introduction and related work}
Face replacement or face swapping is relevant in many scenarios including
the provision of privacy, appearance transfiguration in portraits, video compositing, and other
creative applications. The exact formulation of this problem varies depending on the application, with
some goals easier to achieve than others.

Bitouk \etal~\cite{bitouk08}, for example, automatically substituted an input face by another face
selected from a large database of images based on the similarity of appearance and pose.
The method replaces the eyes, nose, and mouth of the face and further makes color and illumination
adjustments in order to blend the two faces. This design has two major limitations which we address in this
paper: there is no control over the output identity and the expression of the input face is altered.


A more difficult problem was addressed by Dale \etal~\cite{dale11}. Their work focused on the
replacement of faces in videos, where video footage of two subjects performing similar roles are available. Compared to static images,
sequential data poses extra difficulties of temporal alignment, tracking facial performance and ensuring temporal consistency
of the resulting footage. The resulting system is complex and still requires a substantial amount of time and
user guidance.

One notable approach trying to solve the related problem of pupeteering -- that is, controlling the expression
of one face with another face -- was presented by Suwajanakorn \etal~\cite{ira15}.
The core idea is to build a 3D model of both the input and the replacement face from a large number of images.
That is, it only works well where a few hundred images are available but cannot be applied to single images.

The abovementioned approaches are based on complex multistage systems combining
algorithms for face reconstruction, tracking, alignment and image compositing.
These systems achieve convincing results which are sometimes indistinguishable from real photographs.
However, none of these fully addresses the problem which we introduce below.

\begin{figure}[t]
    \begin{center}
      \subcaptionbox{\label{fig:teaser_a}}{\includegraphics[width =0.3\linewidth]{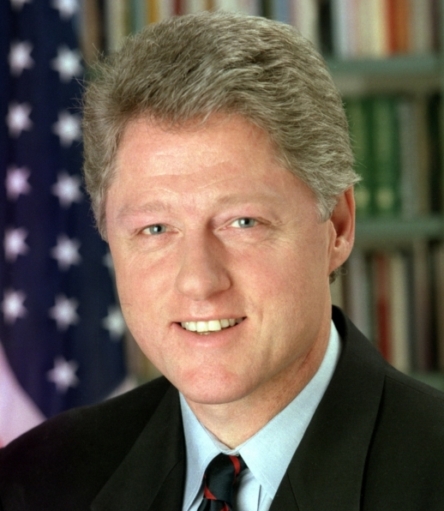}}
      \subcaptionbox{\label{fig:teaser_b}}{\includegraphics[width =0.3\linewidth]{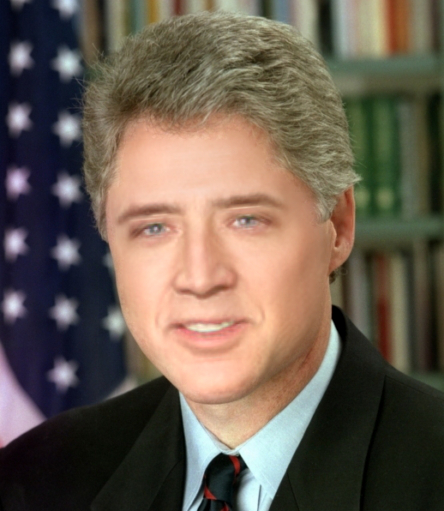}}
      \subcaptionbox{\label{fig:teaser_c}}{\includegraphics[width =0.3\linewidth]{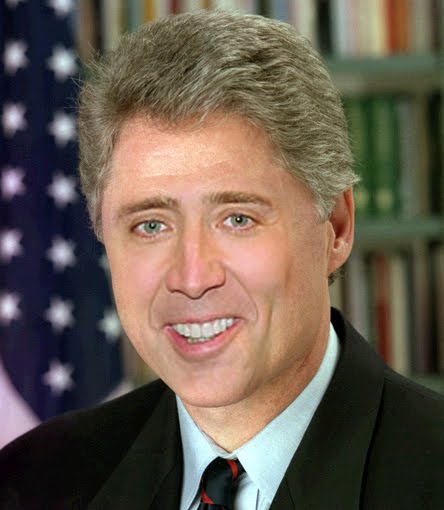}}
    \end{center}
    \vskip -0.2in
    \caption{(a)~The input image. (b)~The result of face swapping with Nicolas Cage using our method.
    (c)~The result of a manual face swap (source: \url{http://niccageaseveryone.blogspot.com}).}
    \vskip -0.2in
    \label{fig:teaser}
\end{figure}

\begin{figure*}[ht!]
    \begin{center}
    \includegraphics[width=0.85\linewidth]{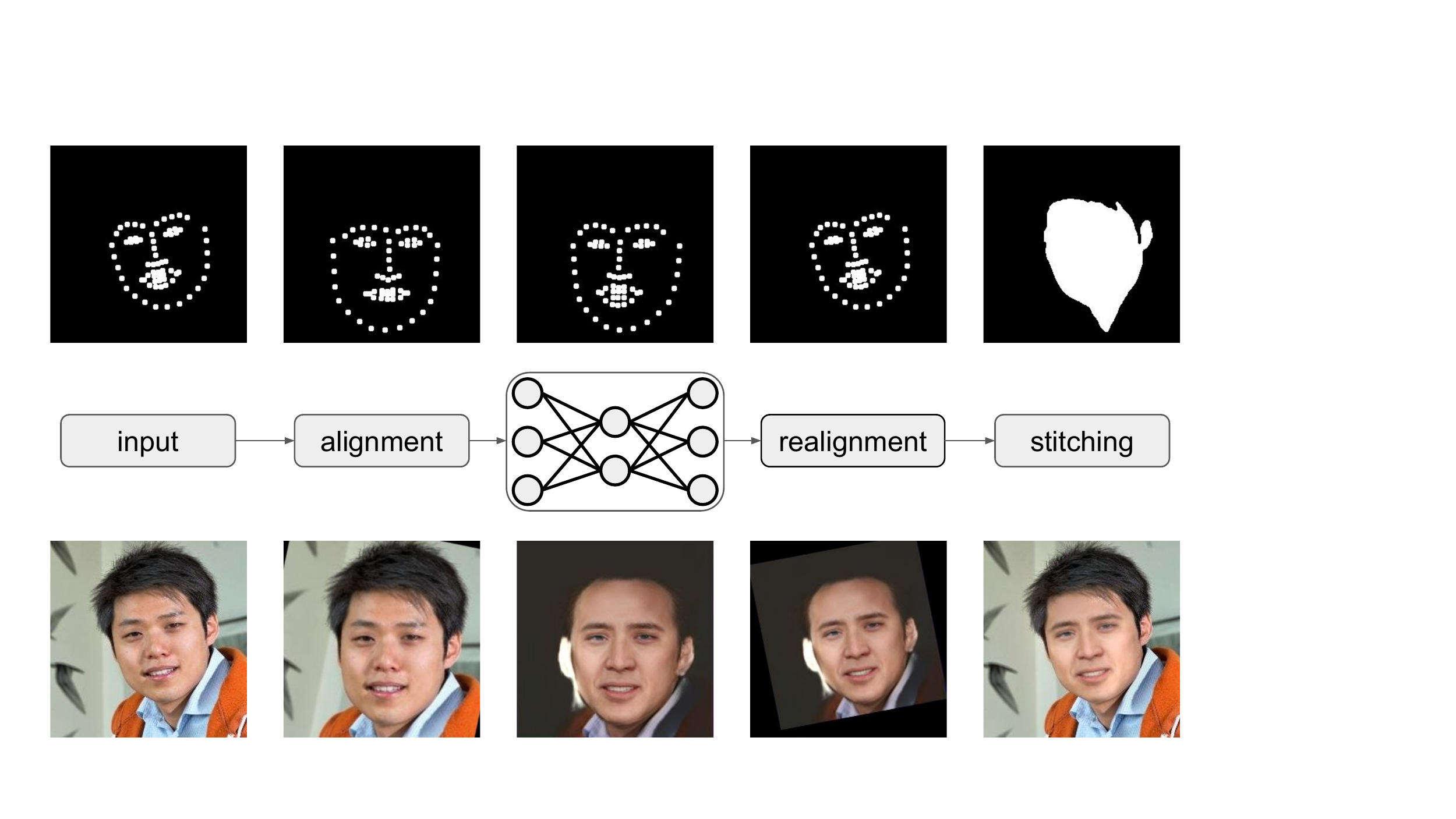}
    \end{center}
    \caption{A schematic illustration of our approach. After aligning the input face to a reference image,
    a convolutional neural network is used to modify it.
    Afterwards, the generated face is realigned and combined with the input image by using a
    segmentation mask. The top row shows facial keypoints used to define the
    affine transformations of the alignment and realignment steps, and the skin segmentation mask
    used for stitching.}
    \label{fig:model}
\end{figure*}

\textbf{Problem outline:} We consider the case where given a single input image of any person A, we
would like to replace his/her identity with that of another person B, while keeping the input pose,
facial expression, gaze direction, hairstyle and lighting intact. An example is given in Figure~\ref{fig:teaser},
where the original identity (Figure~\ref{fig:teaser_a}) was altered with little or no changes to the other factors
(Figure~\ref{fig:teaser_b}).

We propose a novel solution which is inspired by recent
progress in artistic style transfer \cite{gatys16,li16},
where the goal is to render the semantic content of one image in the style of
another image. The foundational work of Gatys \etal~\cite{gatys16} defines the concepts of
content and style as functions in the feature space of convolutional neural networks trained for
object recognition. Stylization is carried out using a rather slow and memory-consuming optimization process.
It gradually changes pixel values of an image until its content
and style statistics match those from a given content image and a given style image, respectively.

An alternative to the expensive optimization approach was proposed by Ulyanov \etal~\cite{ulyanov16} and Johnson \etal~\cite{johnson16}.
They trained feed-forward neural networks to transform any image into its stylized version,
thus moving costly computations to the training stage of the network.
At test time, stylization requires a single forward pass through the network,
which can be done in real time. The price of this improvement is that a separate network has
to be trained per style.

\begin{figure*}[ht!]
    \begin{center}
        \includegraphics[width=\linewidth]{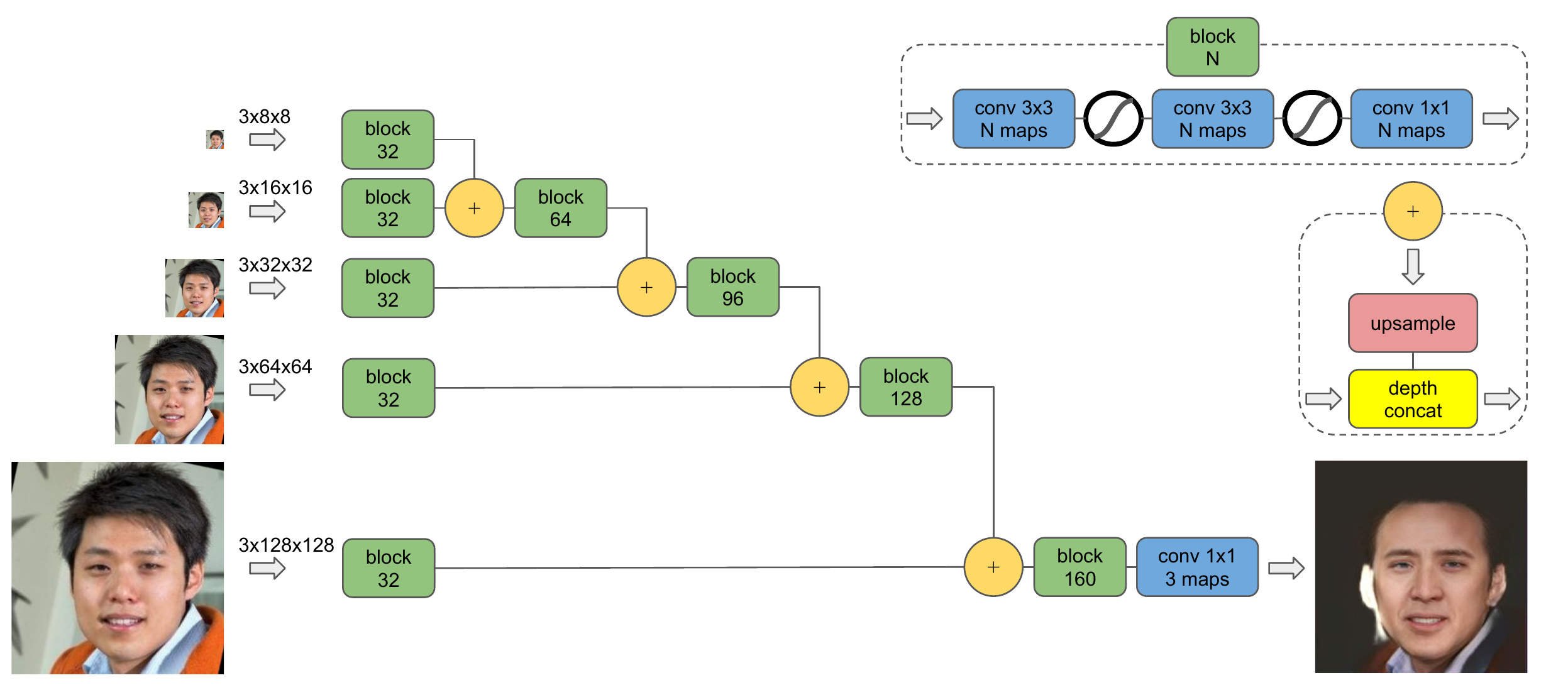}
    \end{center}
    \caption{Following Ulyanov \etal~\cite{ulyanov16}, our transformation network has a multi-scale
    architecture with inputs at different resolutions.}
    \label{fig:genmodel}
\end{figure*}

While achieving remarkable results on transferring the style of many artworks, the neural style transfer method
is less suited for photorealistic transfer. The reason appears to be that the Gram matrices used to represent the style
do not capture enough information about the spatial layout of the image.
This introduces unnatural distortions which go unnoticed in artistic images but not in real images.
Li and Wand~\cite{li16} alleviated this problem
by replacing the correlation-based style loss with a patch-based loss preserving the local structures better.
Their results were the first to suggest that photo-realistic and controlled modifications of photographs of faces
may be possible using style transfer techniques. However, this direction was left fairly unexplored and
like the work of Gatys \etal~\cite{gatys16}, the approach depended on expensive
optimization. Later applications of the patch-based loss to feed-forward neural networks only explored texture
synthesis and artistic style transfer~\cite{li16b}.

This paper takes a step forward upon the work of Li and Wand~\cite{li16}:
we present a feed-forward neural network, which achieves high levels of photorealism in generated face-swapped images.
The key component is that our method, unlike previous approaches to style transfer, uses a multi-image style loss,
thus approximating a manifold describing a style rather than using a single reference point.
We furthermore extend the loss function to explicitly match lighting conditions between images.
Notably, the trained networks allow us to perform face swapping in, or near, real time.
The main requirement for our method is to have a collection of images from the target (replacement) identity.
For well photographed people whose images are available on the Internet, this collection can be easily obtained.

Since our approach to face replacement is rather unique, the results look different from
those obtained with more classical computer vision techniques~\cite{bitouk08,dale11,ira16}
or using image editing software (compare Figures~\ref{fig:teaser_b} and~\ref{fig:teaser_c}).
While it is difficult to compete with an artist specializing in this task, our results suggest
that achieving human-level performance may be possible with a fast and automated approach.

\section{Method}

Having an image of person A, we would like to transform his/her identity into person B's identity
while keeping head pose and expression as well as lighting conditions intact.
In terms of style transfer, we think of input image A's pose and expression as the
\textit{content}, and input image B's identity as the \textit{style}. Light is dealt with
in a separate way introduced below.

Following Ulyanov \etal~\cite{ulyanov16} and Johnson \etal~\cite{johnson16}, we use a convolutional neural network parameterized
by weights $\mathbf{W}$ to transform the content image $\mathbf{x}$, i.e. input image A, into the output
image $\mathbf{\hat{x}} = f_\mathbf{W}(\mathbf{x})$. Unlike previous work, we assume that we are
given not one but a set of style images which we denote by $\mathbf{Y} = \{ \mathbf{y}_1, \dots,
\mathbf{y}_N \}$. These images describe the identity which we would like to match and are only used
during training of the network.

Our system has two additional components performing face alignment and background/hair/skin segmentation.
We assume that all images (content and style), are aligned to a frontal-view reference face.
This is achieved using an affine transformation, which aligns 68 facial keypoints from
a given image to the reference keypoints. Facial keypoints were extracted using \textit{dlib}~\cite{dlib}.
Segmentation is used to restore the background and hair of the input image~$\mathbf{x}$,
which is currently not preserved by our transformation network. We used a seamless cloning technique~\cite{seamless_clone}
available in OpenCV~\cite{opencv} to stitch the background and the resulting face-swapped image.
While fast and relatively accurate methods for segmentation exist, including some based on neural
networks~\cite{bansal:2016,Long:2015,Paszke:2016}, we assume for simplicity that a
segmentation mask is given and focus on the remaining problems. An overview of the system is given
in Figure~\ref{fig:model}.

In the following we will describe the architecture of the transformation network and the loss functions
used for its training.

\subsection{Transformation network}
\label{sec:architecture}
The architecture of our transformation network is based on the architecture of Ulyanov \etal~\cite{ulyanov16}
and is shown in Figure~\ref{fig:genmodel}. It is a multiscale architecture with branches operating on different
downsampled versions of the input image~$\mathbf{x}$. Each such branch has blocks of zero-padded convolutions followed by
linear rectification. Branches are combined via nearest-neighbor upsampling by a factor of two and concatenation along the channel axis.
The last branch of the network ends with a $1\times1$ convolution and 3 color channels.

The network in Figure~\ref{fig:genmodel}, which is designed for $128\times 128$ inputs, has 1M parameters.
For larger inputs, e.g. $256\times256$ or $512\times512$, it is straightforward to infer the architecture
of the extra branches. The network output is obtained only from the branch with the highest resolution.

We found it convenient to firstly train the network on $128\times 128$ inputs,
and then use it as a starting point for the network operating on larger images.
In this way, we can achieve higher resolutions without the need to retrain the whole model.
Although, we are restrained by the availability of high quality image data for model's training.

\subsection{Loss functions}

For every input image $\mathbf{x}$, we aim to generate an $\mathbf{\hat x}$ which jointly minimizes
the following content and style loss.
These losses are defined in the feature space of the normalised version of the 19-layer VGG network~\cite{gatys16,vgg}.
We will denote the VGG representation of $\mathbf{x}$ on layer $l$ as $\Phi_l(\mathbf{x})$.
Here we assume that $\mathbf{x}$ and every style image $\mathbf{y}$ are aligned to a reference face.
All images have the dimensionality of $3 \times H \times W$.

\vspace{1mm}
\noindent\textbf{Content loss:}
For the $l$th layer of the VGG network, the content loss is given by \cite{gatys16}:

\begin{equation}
    \mathcal{L}_{content}(\mathbf{\hat{x}}, \mathbf{x}, l) = \frac1{\left\vert{\Phi_l(\mathbf{x})}\right\vert}\|\Phi_l(\mathbf{\hat x}) - \Phi_l(\mathbf{x})\|_2^2,
    \label{eq:l_content_loss}
\end{equation}

where $\left\vert{\Phi_l(\mathbf{x})}\right\vert = C_l H_l W_l$ is the dimensionality of $\Phi_l(\mathbf{x})$ with shape
$C_l \times H_l \times W_l$.

In general, the content loss can be computed over multiple layers of the network, so that the overall content loss would be:
\begin{equation}
    \mathcal{L}_{content}(\mathbf{\hat{x}}, \mathbf{x}) = \sum_{l} \mathcal{L}_{content}(\mathbf{\hat{x}}, \mathbf{x}, l)
    \label{eq:content_loss}
\end{equation}

\vspace{1mm}
\noindent\textbf{Style loss:}
Our loss function is inspired by the patch-based loss of Li and Wand~\cite{li16}.
Following their notation, let $\Psi(\Phi_l(\mathbf{\hat x}))$ denote the list of all patches
generated by looping over $H_l \times W_l$ possible locations in $\Phi_l(\mathbf{\hat x})$
and extracting a squared $k \times k$ neighbourhood around each point.
This process yields $M = (H_l-k+1) \times (W_l-k+1)$ neural patches, where the $i$th patch $\Psi_i(\Phi_l(\mathbf{\hat{x}}))$
has dimensions of $C_l \times k \times k$.

For every such patch from $\mathbf{\hat x}$ we find the best matching patch among patches extracted from $\mathbf{Y}$
and minimize the distance between them. As an error metric we used the cosine distance $d_c$:
\begin{align}
    d_c(\mathbf{u}, \mathbf{v})
    &= 1 - \frac{\mathbf{u}^\top \mathbf{v}}{||\mathbf{u}|| \cdot ||\mathbf{v}||}, \\
    \mathcal{L}_{style}(\mathbf{\hat{x}}, \mathbf{Y}, l)
    &= \frac1{M} \sum_{i = 1}^{M} d_c\left( \Psi_i(\Phi_l(\mathbf{\hat{x}})), \Psi_i(\Phi_l(\mathbf{y}_{NN(i)})) \right),
    \label{eq:l_style_loss}
\end{align}
where $NN(i)$ selects for each patch a corresponding style image.
Unlike Li and Wand~\cite{li16}, who used a single style image $\mathbf{y}$ and selected a patch among
all possible patches $\Psi(\Phi_l(\mathbf{y}))$, we only search for patches in the same location~$i$,
but across multiple style images:
\begin{equation}
    NN(i) = \argmin{j = 1,...,N_{best}} d_c\left( \Psi_i(\Phi_l(\mathbf{\hat{x}})), \Psi_i(\Phi_l(\mathbf{y}_j)) \right).
    \label{eq:nn}
\end{equation}
We found that only taking the best matching $N_{best} < N$ style images into account worked better, which here are assumed
to be sorted according to the Euclidean distance between their
facial landmarks and landmarks of the input image $\mathbf{x}$. In this way every training image has a costumized set of
style images, namely those with similar pose and expression.

Similar to Equation~\ref{eq:content_loss}, we can compute style loss over multiple layers of the VGG.

\begin{figure}
\begin{center}
\includegraphics[width=\linewidth]{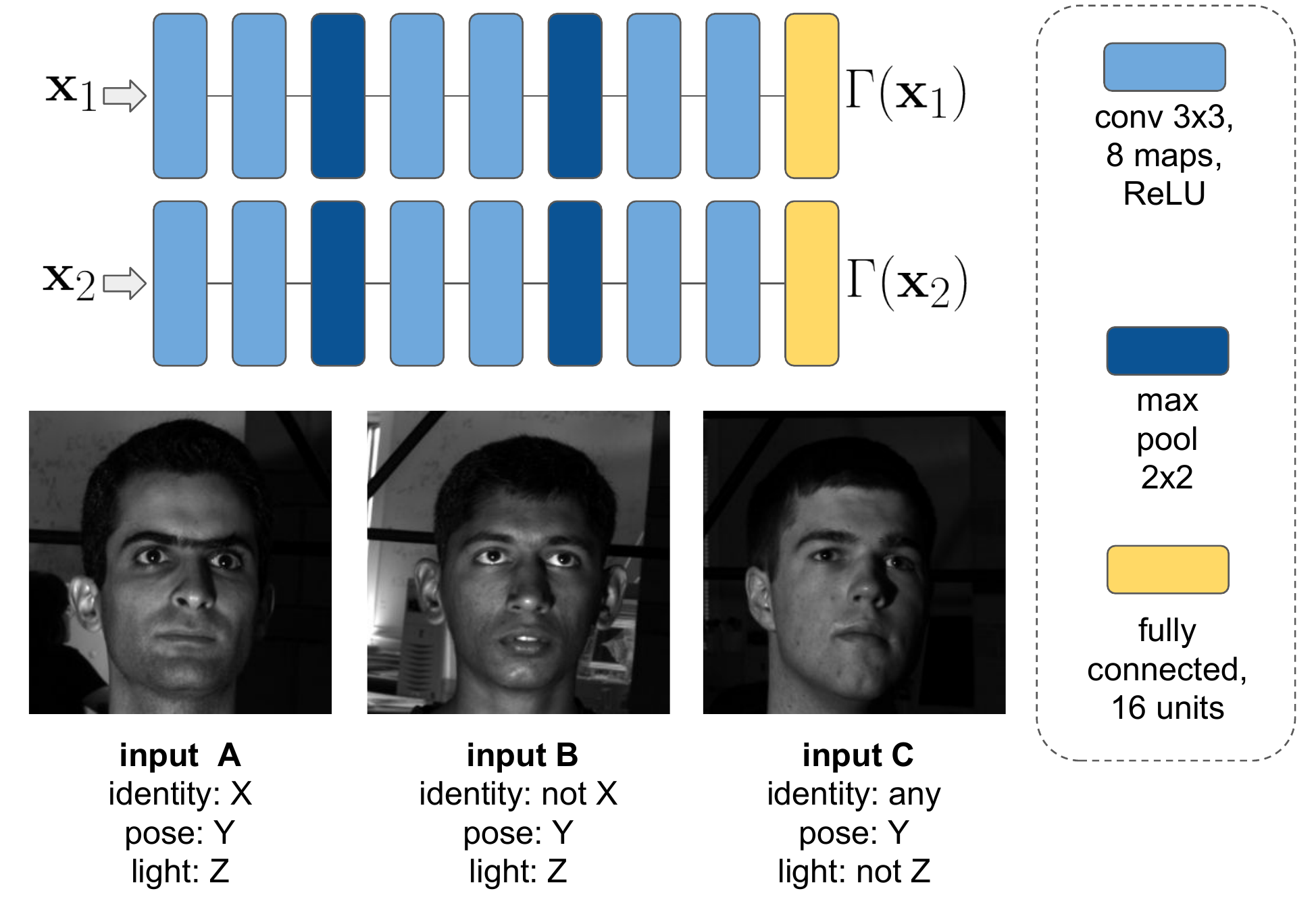}
\end{center}
   \caption{The lighting network is a siamese network
   trained to maximize the distance between images with different lighting conditions
   (inputs A and C) and to minimize this distance for pairs with equal illumination (inputs A and B).
   The distance is defined as an $L_2$ norm in the feature space of the fully connected layer.
   All input images are aligned to the same reference face as for the inputs to the transformation network.}
   \label{fig:light_model}
\end{figure}

\begin{figure*}
\centering
\begin{subfigure}[b]{\textwidth}
   \includegraphics[width=1\linewidth]{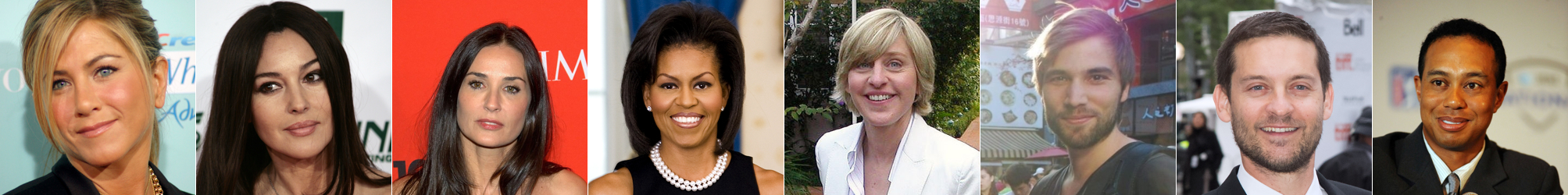}
   \caption{}
   \label{fig:lwf_a}
\end{subfigure}
\begin{subfigure}[b]{\textwidth}
   \includegraphics[width=1\linewidth]{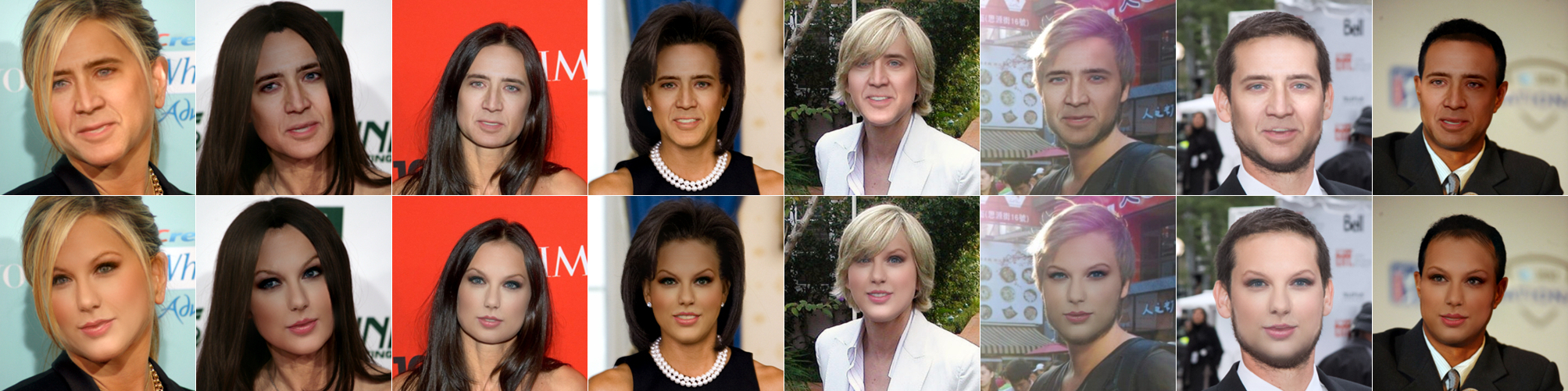}
   \caption{}
   \label{fig:lwf_b}
\end{subfigure}
\begin{subfigure}[b]{\textwidth}
   \includegraphics[width=1\linewidth]{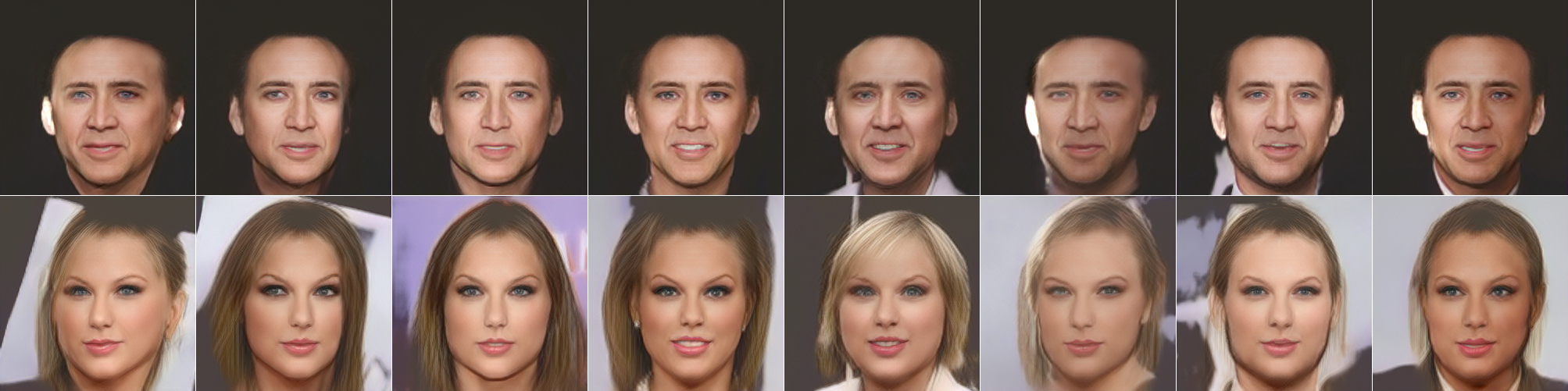}
   \caption{}
   \label{fig:lwf_c}
\end{subfigure}

\caption[]{(a) Original images. (b) Top: results of face swapping with Nicolas Cage, bottom:
   results of face swapping with Taylor Swift. (c) Top: raw outputs of CageNet, bottom: outputs of
   SwiftNet. Note how our method alters the appearance of the
   nose, eyes, eyebrows, lips and facial wrinkles. It keeps gaze direction, pose and lip
   expression intact, but in a way which is natural for the target identity. Images are best viewed electronically.}
\end{figure*}

\vspace{1mm}
\noindent\textbf{Light loss:}
Unfortunately, the lighting conditions of the content image $\mathbf{x}$ are not preserved in the generated image $\mathbf{\hat x}$
when only using the above-mentioned losses defined in the VGG's feature space.
We address this problem by introducing an extra term to our objective which penalizes changes in illumination.
To define the illumination penalty, we exploited the idea of using a feature space of a pretrained network
in the same way as we used VGG for the style and content. Such an approach would work if the feature space
represented differences in lighting conditions. The VGG network is not appropriate
for this task since it was trained for classifying objects, where illumination information is not particularly relevant.

To get the desirable property of lighting sensitivity, we constructed a small siamese convolutional neural network~\cite{siamese}.
It was trained to discriminate between pairs of images with either equal or different illumination conditions.
Pairs of images always had equal pose.
We used the Exteded Yale Face Database B~\cite{yale}, which contains grayscale portraits of subjects under 9
poses and 64 lighting conditions. The architecture of the lighting network
is shown in Figure~\ref{fig:light_model}.

We will denote the feature representation of $\mathbf{x}$ in the last layer of the lighting network as $\Gamma(\mathbf{x})$
and introduce the following loss function, which tries to prevent generated images $\mathbf{\hat x}$ from
having different illumination conditions than those from the content image $\mathbf{x}$. Both $\mathbf{\hat x}$
and $\mathbf{x}$ are single-channel luminance images.

\begin{equation}
    \mathcal{L}_{light}(\mathbf{\hat{x}}, \mathbf{x})
    = \frac1{\left\vert{\Gamma(\mathbf{x})}\right\vert}\|\Gamma(\mathbf{\hat x}) - \Gamma(\mathbf{x})\|_2^2
    \label{eq:light_loss}
\end{equation}

\vspace{1mm}
\noindent\textbf{Total variation regularization:}
Following the work of Johnson~\cite{johnson16} and others, we used regularization to encourage spatial smoothness:

\begin{equation}
    \mathcal{L}_{TV}(\mathbf{\hat{x}}) = \sum_{i, j}(\mathbf{\hat x}_{i, j + 1} - \mathbf{\hat x}_{i, j})^{2} + (\mathbf{\hat x}_{i + 1, j} - \mathbf{\hat x}_{i, j})^{2}
    \label{eq:tv_loss}
\end{equation}

The final loss function is a weighted combination of the described losses:
\begin{equation}
    \begin{aligned}
        \mathcal{L}(\mathbf{\hat{x}},\mathbf{x},\mathbf{Y}) =\, & \mathcal{L}_{content}(\mathbf{\hat{x}}, \mathbf{x})
        + \alpha \mathcal{L}_{style}(\mathbf{\hat{x}}, \mathbf{Y}) + \\
                                              & \beta \mathcal{L}_{light}(\mathbf{\hat{x}}, \mathbf{x}) + \gamma \mathcal{L}_{TV}(\mathbf{\hat{x}})
    \end{aligned}
    \label{eq:total_loss}
\end{equation}

\section{Experiments}

\subsection{CageNet and SwiftNet}

\vspace{1mm}
\noindent\textbf{Technical details:}
We trained a transformation network to perform the face swapping with Nicolas Cage,
of whom we collected about 60 photos from the Internet with different poses and facial expressions.
To further increase the number of style images, every image was horizontally flipped.
As a source of content images for training we used the CelebA dataset~\cite{celeba}, which contains
over 200,000 images of celebrities.

Training of the network was performed in two stages.
Firstly, the network described in Section~\ref{sec:architecture} was trained to process
$128 \times 128$ images. It minimized the objective function given by Equation~\ref{eq:total_loss},
where $\mathcal{L}_{light}$ was computed using a lighting network also trained on $128\times128$ inputs.
In Equation~\ref{eq:total_loss}, we used $\beta=10^{-22}$ to make the lighting loss
$\mathcal{L}_{light}$ comparable to content and style losses. For the total variation loss,
we chose $\gamma=0.3$ .

Training the transformation network with Adam~\cite{adam} for 10K iterations with a batch size of 16
took 2.5 hours on a Tesla M40 GPU (Theano~\cite{theano} and Lasagne~\cite{lasagne} implementation).
Weights were initialized orthogonally~\cite{saxe13}. The learning rate was
decreased from 0.001 to 0.0001 over the course of the training following a manual learning rate
schedule.

\begin{figure}[t]
\begin{center}
   \includegraphics[width=0.95\linewidth]{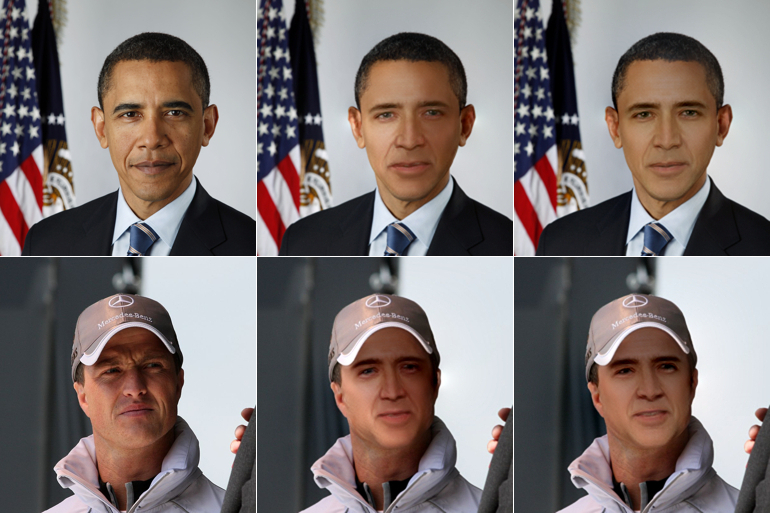}
\end{center}
   \caption{Left: original image, middle and right: CageNet trained with
   and without the lighting loss.}
\label{fig:light}
\end{figure}

With regards to the specifics of style transfer, we used the following settings.
Style losses and content loss were computed using VGG layers $\{\verb.relu3_1., \verb.relu4_1.\}$ and $\{\verb.relu4_2.\}$
respectively. For the style loss, we used a patch size of $k=1$. During training, each input image was matched
to a set of $N_{best}$ style images, where $N_{best}$ was equal to 16.
The style weight $\alpha$ in the total objective function (Equation~\ref{eq:total_loss})
was the most crucial parameter to tune. Starting from $\alpha = 0$ and gradually increasing
it to $\alpha=20$ yielded the best results in our experiments.

Having trained a model for $128 \times 128$ inputs and outputs,
we added an extra branch for processing $256 \times 256$ images.
The additional branch was optimized while keeping the rest of the network fixed.
The training protocol for this network was identical
to the one described above, except the style weight $\alpha$ was increased to 80 and
we used the lighting network trained on $256\times256$ inputs.
The transformation network takes 12 hours to train and has about 2M parameters,
of which half are trained during the second stage.

\vspace{1mm}
\noindent\textbf{Results:}
Figure~\ref{fig:lwf_b} shows the final results of our face swapping method applied to a selection of images
in Figure~\ref{fig:lwf_a}. The raw outputs of the neural network are given in Figure~\ref{fig:lwf_c}.
We find that the neural network is able to introduce noticeable changes to the appearance of a face while
keeping head pose, facial expression and lighting intact. Notably, it significantly alters the
appearance of the nose, eyes, eyebrows, lips, and wrinkles in the faces, while keeping gaze
direction and still producing a plausible image. However, coarser features such as the overall head
shape are mostly unaltered by our approach, which in some cases diminishes the effect of a perceived
change in identity. One can notice that when target and input identities have different skin colors,
the resulting face has an average skin tone. This is partly due to the seamless cloning of the swapped image
with the background, and to a certain extent due to the transformation network.
The latter fuses the colors because its loss function is based on the VGG network, which is color sensitive.

To test how our results generalize to other identities, we trained the same transformation network
using approximately 60 images of Taylor Swift. We find that results of similar quality can be achieved
with the same hyperparameters (Figure~\ref{fig:lwf_b}).

\begin{figure}[t]
\begin{center}
 \includegraphics[width=0.95\linewidth]{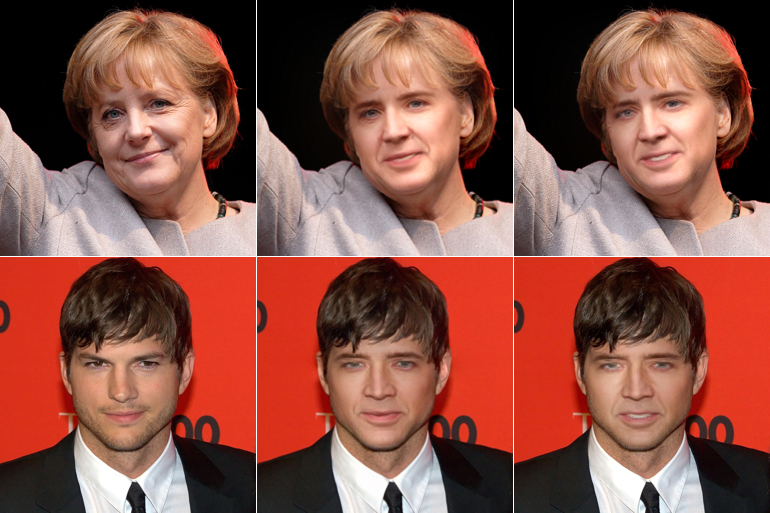}
\end{center}
   \caption{Left: original image, middle and right: CageNet trained on $256\times 256$ images
   with style weights $\alpha=80$ and $\alpha=120$ respectively. Note how facial expression is
   altered in the latter case.}
\label{fig:high_loss}
\end{figure}

\begin{figure}[t]
\begin{center}
 \includegraphics[width=0.95\linewidth]{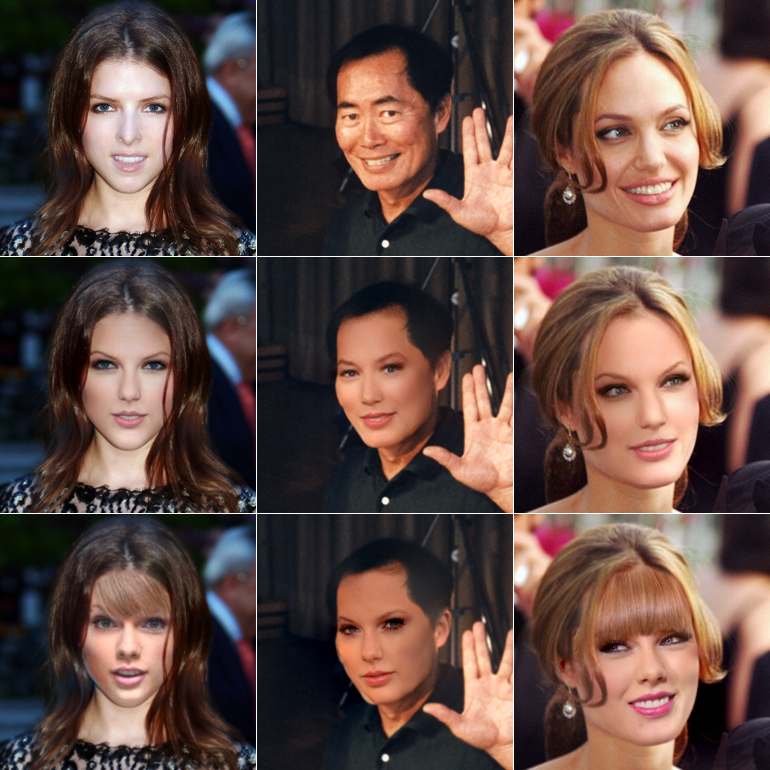}
\end{center}
   \caption{Top: original images, middle: results of face swapping with Taylor Swift
   using our method, bottom: results of a baseline approach.
   Note how the baseline method changes facial expressions,
   gaze direction and the face does not always blend in well with the surrounding image.
   }
\label{fig:baseline}
\end{figure}

Figure~\ref{fig:light} shows the effect of the lighting loss in the total
objective function. When no such loss is included, images generated with CageNet
have flat lighting and lack shadows.

While the generated faces often clearly look like the target identity,
it is in some cases difficult to recognize the person because features
of the input identity remain in the output image.
They could be completely eliminated by increasing the weight
of the style loss. However, this comes at the cost of ignoring
the input's facial expression as shown in Figure~\ref{fig:high_loss},
which we do not consider to be a desirable behaviour since it
changes the underlying emotional interpretation of the image.
Indeed, the ability to transfer expressions distinguishes our approach
from other methods operating on a single image input.
To make the comparison clear, we implemented a simple face swapping method
which performs the same steps as in Figure~\ref{fig:model},
except for the application of the transformation network.
This step was replaced by selecting an image from the style set
whose facial landmarks are closest to those from the input image.
The results are shown in Figure~\ref{fig:baseline}.
While the baseline method trivially produces sharp looking faces,
it alters expressions, gaze direction and faces generally blend in worse with the rest of the image.

\begin{figure*}[t]
\begin{center}
   \includegraphics[width=\linewidth]{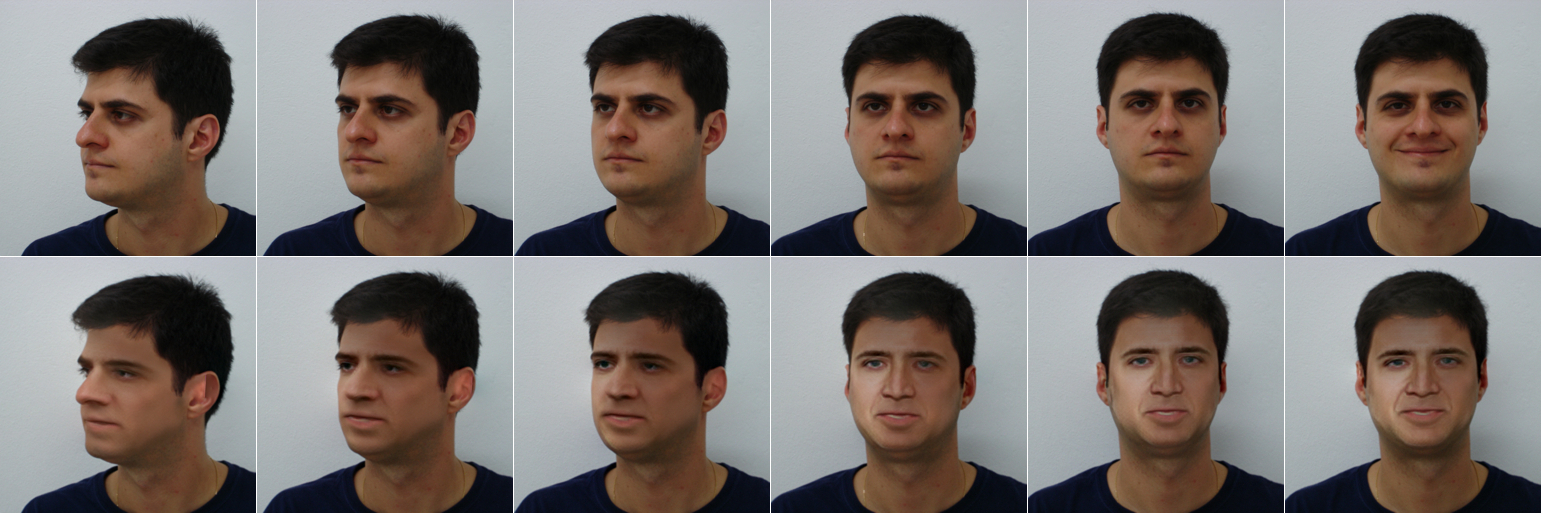}
\end{center}
   \caption{Top: original images, bottom: results of face swapping with Nicolas Cage. Note how the
   identity of Nicolas Cage becomes more identifiable as the view changes from side to frontal. Also note
   how the light is wrongly amplified in some of the images.}
\label{fig:fei}
\end{figure*}

In the following, we explore a few failure cases of our approach.
We noticed that our network works better for frontal views than for profile views. In Figure~\ref{fig:fei}
we see that as we progress from the side view to the frontal view, the face becomes more recognizable as Nicolas Cage.
This may be caused by an imbalance in the datasets. Both our training set (CelebA) and the set of style images included a lot more frontal views than profile views
due to the prevalence of these images on the Internet. Figure~\ref{fig:fei} also illustrates the failure
of the illumination transfer where the network amplifies the sidelights. The reason might be the prevalence of
images with harsh illumination conditions in the training dataset of the lighting network.

\begin{figure}[t]
	\begin{center}
	\includegraphics[width=0.95\linewidth]{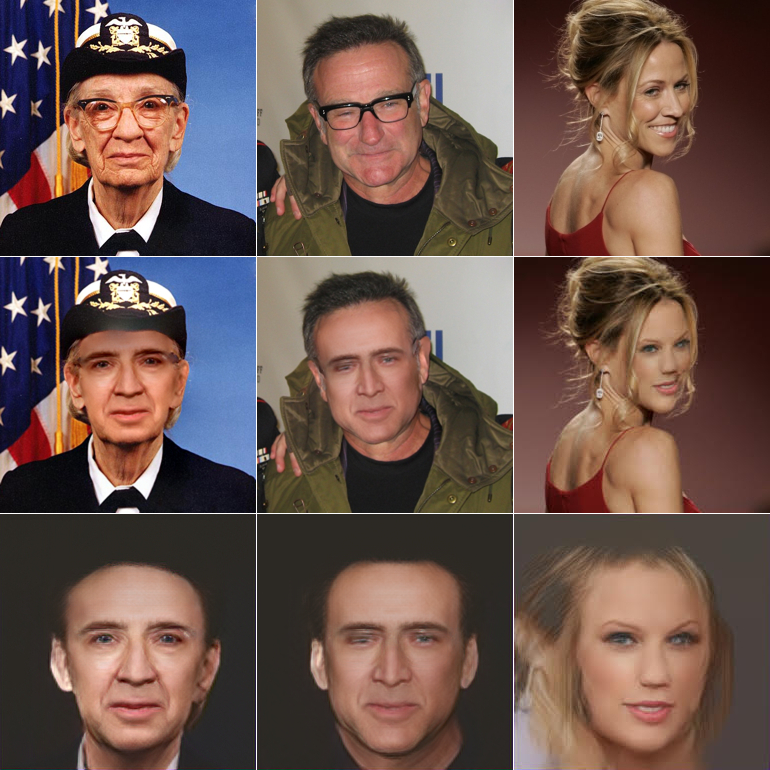}
	\end{center}
	\caption{Examples of problematic cases.
		Left and middle: facial occlusions, in this case glasses, are not preserved and can lead to
		artefacts. Middle: closed eyes are not swapped correctly, since no image in the style set had this expression.
		Right: poor quality due to a difficult pose, expression, and hair style.}
	\label{fig:fail}
\end{figure}

Figure~\ref{fig:fail} demonstrates other examples which are currently not handled well by our
approach. In particular, occluding objects such as glasses are removed by the network and can lead
to artefacts.

\vspace{1mm}
\noindent\textbf{Speed and Memory:}
A feed-forward pass through the transformation network takes 40\,ms for a $256 \times 256$
input image on a GTX Titan X GPU. For the results presented in this paper, we manually segmented images
into skin and background regions. However, a simple network we trained for automatic
segmentation~\cite{unet}, can produce reasonable masks in about 5\,ms.
Approximately the same amount of CPU time (i7-5500U) is needed for image
alignment. While we used $dlib$~\cite{dlib} for facial keypoints detection, much faster methods
exist which can run in less than 0.1\,ms ~\cite{Ren:2014}. Seamless cloning using $OpenCV$ on
average takes 35\,ms.

At test time, style images do not have to be supplied to the network,
so the memory consumption is low.

\section{Discussion and future work}

By the nature of style transfer, it is not feasible to evaluate our results quantitatively based on the values of
the loss function~\cite{gatys16}. Therefore, our analysis was limited to subjective evaluation only.
The departure of our approach from conventional practices in face swapping makes it difficult to perform
a fair comparison to prior works. Methods, which solely manipulate images~\cite{bitouk08,ira16} are capable of
producing very crisp images, but they are not able to transfer facial poses and expressions accurately
given a limited number of photographs from the target identity. More complex approaches,
on the other hand, require many images from the person we want to replace~\cite{dale11,ira15}.

Compared to previous style transfer results our method achieves high levels of photorealism.
However, they can still be improved in multiple ways. Firstly, the quality of generated results depends on the collection of style images.
Face replacement of a frontal view typically results in better quality compared to profile views.
This is likely due to a greater number of frontal view portraits found on the Internet.
Another source of problems are uncommon facial expressions and harsh lighting conditions from the input to the face swapped image.
It may be possible to reduce these problems with larger and more carefully chosen photo collections.
Some images also appear oversmoothed. This may be improved in future work by adding an adversarial
loss, which has been shown to work well in combination with VGG-based losses~\cite{Ledig:2016,Sonderby:2016}.

Another potential improvement would be to modify the loss function so that the transformation network preserves
occluding objects such as glasses. Similarly, we can try to penalize the network for changing the
background of the input image. Here we used segmentation in a post-processing step to preserve the background.
This could be automated by combining our network with a neural network trained for segmentation~\cite{liu15, unet}.

Further improvements may be achieved by enhancing the facial keypoint detection algorithm.
In this work, we used $dlib$~\cite{dlib}, which is accurate only up to a certain degree of head rotation.
For extreme angles of view, the algorithm tries to approximate the location of invisible keypoints
by fitting an average frontal face shape. Usually this results in inaccuracies for points along the jawline, which
cause artifacts in the resulting face-swapped images.

Other small gains may be possible when using the \textit{VGG-Face}~\cite{vgg_face} network for the content and style loss
as suggested by Li~\etal~\cite{attr_net}.
Unlike the VGG network used here, which was trained to classify images from various categories~\cite{imagenet}, VGG-Face
was trained to recognize about 3K unique individuals. Therefore, the feature space of VGG-Face
would likely be more suitable for our problem.

\section{Conclusion}

In this paper we provided a proof of concept for a fully-automatic
nearly real-time face swap with deep neural networks. We introduced a new objective and showed that style transfer
using neural networks can generate realistic images of human faces. The proposed method deals with a specific type of face
replacement. Here, the main difficulty was to change the identity without altering the original pose,
facial expression and lighting. To the best of our knowledge, this particular problem has not been addressed previously.

While there are certainly still some issues to overcome, we feel we made significant progress on the
challenging problem of neural-network based face swapping. There are many advantages to using feed-forward
neural networks, e.g., ease of implementation, ease of adding new identities, ability to control the strength of
the effect, or the potential to achieve much more natural looking results.

\section*{Photo credits}

The copyright of the photograph used in Figure~\ref{fig:model} is owned by Peter Matthews.
Other photographs were part of the public domain or made available under
a CC license by the following rights holders: Angela George, Manfred Werner, David Shankbone,
Alan Light, Gordon Correll, AngMoKio, Aleph, Diane Krauss, Georges Biard.

{\small
\bibliographystyle{ieee}
\bibliography{egbib}

\begin{thebibliography}{10}\itemsep=-1pt

\bibitem{bansal:2016}
A.~Bansal, X.~Chen, B.~Russell, A.~Gupta, and D.~Ramanan.
\newblock {PixelNet: Towards a General Pixel-Level Architecture}, 2016.
\newblock arXiv:1609.06694v1.

\bibitem{bitouk08}
D.~Bitouk, N.~Kumar, S.~Dhillon, P.~Belhumeur, and S.~K. Nayar.
\newblock Face swapping: Automatically replacing faces in photographs.
\newblock In {\em ACM Transactions on Graphics (SIGGRAPH)}, 2008.

\bibitem{siamese}
S.~Chopra, R.~Hadsell, and Y.~Lecun.
\newblock Learning a similarity metric discriminatively, with application to
  face verification.
\newblock In {\em IEEE Conference on Computer Vision and Pattern Recognition
  (CVPR)}, pages 539--546. IEEE Press, 2005.

\bibitem{dale11}
K.~Dale, K.~Sunkavalli, M.~K. Johnson, D.~Vlasic, W.~Matusik, and H.~Pfister.
\newblock Video face replacement.
\newblock {\em ACM Transactions on Graphics (SIGGRAPH)}, 30, 2011.

\bibitem{imagenet}
J.~Deng, W.~Dong, R.~Socher, L.-J. Li, K.~Li, and L.~Fei-Fei.
\newblock {ImageNet: A Large-Scale Hierarchical Image Database}.
\newblock In {\em IEEE Conference on Computer Vision and Pattern Recognition
  (CVPR)}, 2009.

\bibitem{lasagne}
S.~Dieleman, J.~Schluter, C.~Raffel, E.~Olson, S.~K. Sonderby, D.~Nouri,
  D.~Maturana, M.~Thoma, E.~Battenberg, J.~Kelly, J.~D. Fauw, M.~Heilman, D.~M.
  de~Almeida, B.~McFee, H.~Weideman, G.~Takacs, P.~de~Rivaz, J.~Crall,
  G.~Sanders, K.~Rasul, C.~Liu, G.~French, and J.~Degrave.
\newblock Lasagne: First release., Aug. 2015.

\bibitem{gatys16}
L.~A. Gatys, A.~S. Ecker, and M.~Bethge.
\newblock Image style transfer using convolutional neural networks.
\newblock In {\em IEEE Conference on Computer Vision and Pattern Recognition
  (CVPR)}, Jun 2016.

\bibitem{yale}
A.~Georghiades, P.~Belhumeur, and D.~Kriegman.
\newblock From few to many: Illumination cone models for face recognition under
  variable lighting and pose.
\newblock {\em IEEE Trans. Pattern Anal. Mach. Intelligence}, 23(6):643--660,
  2001.

\bibitem{johnson16}
J.~Johnson, A.~Alahi, and L.~Fei{-}Fei.
\newblock Perceptual losses for real-time style transfer and super-resolution.
\newblock In {\em Computer Vision - {ECCV} 2016 - 14th European Conference,
  Amsterdam, The Netherlands, October 11-14, 2016, Proceedings, Part {II}},
  pages 694--711, 2016.

\bibitem{ira16}
I.~Kemelmacher-Shlizerman.
\newblock Transfiguring portraits.
\newblock {\em ACM Transaction on Graphics}, 35(4):94:1--94:8, July 2016.

\bibitem{dlib}
D.~E. King.
\newblock {Dlib-ml: A Machine Learning Toolkit}.
\newblock {\em Journal of Machine Learning Research}, 10:1755--1758, 2009.

\bibitem{adam}
D.~P. Kingma and J.~Ba.
\newblock Adam: {A} method for stochastic optimization, 2014.
\newblock arXiv:1412.6980.

\bibitem{Ledig:2016}
C.~Ledig, L.~Theis, F.~Huszar, J.~Caballero, A.~Aitken, A.~Tejani, J.~Totz,
  Z.~Wang, and W.~Shi.
\newblock {Photo-Realistic Single Image Super-Resolution Using a Generative
  Adversarial Network}, 2016.
\newblock arXiv:1609.04802.

\bibitem{li16}
C.~Li and M.~Wand.
\newblock Combining markov random fields and convolutional neural networks for
  image synthesis.
\newblock In {\em IEEE Conference on Computer Vision and Pattern Recognition
  (CVPR)}, June 2016.

\bibitem{li16b}
C.~Li and M.~Wand.
\newblock Precomputed real-time texture synthesis with markovian generative
  adversarial networks, 2016.
\newblock arXiv:1604.04382v1.

\bibitem{attr_net}
M.~Li, W.~Zuo, and D.~Zhang.
\newblock Convolutional network for attribute-driven and identity-preserving
  human face generation, 2016.
\newblock arXiv:1608.06434.

\bibitem{liu15}
S.~Liu, J.~Yang, C.~Huang, and M.~Yang.
\newblock Multi-objective convolutional learning for face labeling.
\newblock In {\em {IEEE} Conference on Computer Vision and Pattern Recognition,
  ({CVPR}) 2015, Boston, MA, USA, June 7-12, 2015}, pages 3451--3459, 2015.

\bibitem{celeba}
Z.~Liu, P.~Luo, X.~Wang, and X.~Tang.
\newblock Deep learning face attributes in the wild.
\newblock In {\em Proceedings of International Conference on Computer Vision
  (ICCV)}, Dec. 2015.

\bibitem{Long:2015}
J.~Long, E.~Shelhamer, and T.~Darrell.
\newblock Fully convolutional networks for semantic segmentation.
\newblock In {\em IEEE Conference on Computer Vision and Pattern Recognition
  (CVPR)}, pages 3431--3440, 2015.

\bibitem{opencv}
OpenCV.
\newblock Open source computer vision library.
\newblock \url{https://github.com/opencv/opencv}, 2016.
\newblock [Online; accessed 24-October-2016].

\bibitem{vgg_face}
O.~M. Parkhi, A.~Vedaldi, and A.~Zisserman.
\newblock Deep face recognition.
\newblock In {\em British Machine Vision Conference}, 2015.

\bibitem{Paszke:2016}
A.~Paszke, A.~Chaurasia, S.~Kim, and E.~Culurciello.
\newblock {ENet: A Deep Neural Network Architecture for Real-Time Semantic
  Segmentation}, 2016.
\newblock arXiv:1606.02147.

\bibitem{seamless_clone}
P.~P{\'e}rez, M.~Gangnet, and A.~Blake.
\newblock Poisson image editing.
\newblock In {\em ACM Transactions on Graphics (SIGGRAPH)}, SIGGRAPH '03, pages
  313--318, New York, NY, USA, 2003. ACM.

\bibitem{Ren:2014}
S.~Ren, X.~Cao, Y.~Wei, and J.~Sun.
\newblock {Face Alignment at 3000 FPS via Regressing Local Binary Features}.
\newblock In {\em IEEE Conference on Computer Vision and Pattern Recognition
  (CVPR)}, pages 1685--1692, 2014.

\bibitem{unet}
O.~Ronneberger, P.~Fischer, and T.~Brox.
\newblock {\em U-Net: Convolutional Networks for Biomedical Image
  Segmentation}, pages 234--241.
\newblock Springer International Publishing, Cham, 2015.

\bibitem{saxe13}
A.~M. Saxe, J.~L. McClelland, and S.~Ganguli.
\newblock Exact solutions to the nonlinear dynamics of learning in deep linear
  neural networks, 2013.
\newblock arXiv:1312.6120.

\bibitem{vgg}
K.~Simonyan and A.~Zisserman.
\newblock Very deep convolutional networks for large-scale image recognition.
\newblock {\em CoRR}, abs/1409.1556, 2014.

\bibitem{Sonderby:2016}
C.~K. S{\o}nderby, J.~Caballero, L.~Theis, W.~Shi, and F.~Husz\'{a}r.
\newblock {Amortised MAP Inference for Image Super-resolution}.
\newblock {\em arXiv preprint arXiv:1610.04490}, 2016.

\bibitem{ira15}
S.~Suwajanakorn, S.~M. Seitz, and I.~Kemelmacher{-}Shlizerman.
\newblock What makes {T}om {H}anks look like {T}om {H}anks.
\newblock In {\em 2015 {IEEE} International Conference on Computer Vision,
  {ICCV} 2015, Santiago, Chile, December 7-13, 2015}, pages 3952--3960, 2015.

\bibitem{theano}
{Theano Development Team}.
\newblock {Theano: A {Python} framework for fast computation of mathematical
  expressions}, May 2016.
\newblock arXiv:1605.02688.

\bibitem{ulyanov16}
D.~Ulyanov, V.~Lebedev, A.~Vedaldi, and V.~Lempitsky.
\newblock Texture networks: Feed-forward synthesis of textures and stylized
  images.
\newblock In {\em {International Conference on Machine Learning} (ICML)}, 2016.

\end{thebibliography}
}

\end{document}